\documentclass[letterpaper, 10 pt, conference]{ieeeconf}  

\IEEEoverridecommandlockouts                              

\usepackage{graphicx}
\usepackage[outdir=./]{epstopdf}
\DeclareGraphicsExtensions{.eps,.jpg}
\usepackage[table]{xcolor}
\usepackage{balance}
\usepackage{flushend}

\usepackage{siunitx}
\usepackage{amsmath}
\usepackage{caption}
\usepackage{subcaption}

\usepackage{amssymb}
\usepackage{algorithm}
\usepackage{algorithmic}

\usepackage{accents}  

\def \poses{X}
\def \landmarks{L}
\def \associations{D}
\def \measurements{Z}

\def \pose{x}
\def \landmark{\ell}

\newcommand{\opt}[1]{{#1}^*}

\def \Xopt{\opt{\poses}}
\def \Lopt{\opt{\landmarks}}
\def \Dopt{\opt{\associations}}

\newcommand{\geometric}[1]{{#1}^\rho}
\newcommand{\semantic}[1]{{#1}^c}

\DeclareMathOperator{\SE}{SE}

\DeclareMathOperator*{\argmax}{argmax}

\usepackage{accents}  

\def \poses{X}
\def \landmarks{L}
\def \associations{D}
\def \measurements{Z}

\def \pose{x}
\def \landmark{\ell}

\def \Xopt{\opt{\poses}}
\def \Lopt{\opt{\landmarks}}
\def \Dopt{\opt{\associations}}

\title{\LARGE \bf
LOSS-SLAM: Lightweight Open-Set Semantic Simultaneous Localization and Mapping 
}

\author{Kurran Singh$^1$, Tim Magoun$^1$, and John Leonard$^1$
\thanks{$^1$K. Singh, T. Magoun and J. Leonard are with the Computer Science and Artificial Intelligence Laboratory (CSAIL) at the Massachusetts Institute of Technology (MIT), 32 Vassar St, Cambridge, MA 02139, USA.
        Corresponding author: Kurran Singh ({\tt\small singhk@mit.edu})}%
}

\begin{document}

\maketitle
\thispagestyle{empty}
\pagestyle{empty}

\begin{abstract}
Enabling robots to understand the world in terms of objects is a critical building block towards higher level autonomy. The success of foundation models in vision has created the ability to segment and identify nearly all objects in the world. However, utilizing such objects to localize the robot and build an open-set semantic map of the world remains an open research question. In this work, a system of identifying, localizing, and encoding objects is tightly coupled with probabilistic graphical models for performing \textit{open-set} semantic simultaneous localization and mapping (SLAM). Results are presented demonstrating that the proposed lightweight object encoding can be used to perform more accurate object-based SLAM than existing open-set methods, closed-set methods, and geometric  methods while incurring a lower computational overhead than existing open-set mapping methods.  
\end{abstract}

\section*{Supplementary Material}
\label{sect:supplemental-material}
Code implementation and data is made available here: https://kurransingh.github.io/open-set-slam/ 

\section{Introduction}
Developing the ability for robots to understand the world around them in terms of semantically meaningful objects is necessary in order for them to perform higher-level autonomous behaviors (``place the mug in the sink''), interact and cooperate with humans, and have a compressed map representation for low-bandwidth communications \cite{Tolstonogov2021} or for long-term navigation. As we seek to enable such behaviors in an increasing number of situations, it has become more and more critical for robots to be able to perform \textit{open-set} mapping, where the system can detect and map objects even if they are out of distribution from the training dataset. Recent advances in the machine learning community have made it possible to detect such objects, but incorporating such objects into a simultaneous localization and mapping (SLAM) framework remains an open question. Furthermore, most existing open-set works focus on dense mapping, which are useful for tasks in small scale environments, but do not scale to larger environments or long term mapping. This work proposes a computationally efficient method for open-set semantic localization and mapping that utilizes self-supervised vision transformer features (DINO) \cite{Caron2021} to augment geometric correspondence matching at the object level. Specifically, the contributions of this work are as follows: 
\begin{enumerate}
    \item A lightweight (sparse) open-set object representation 

\begin{figure}[h] 
\center
  \includegraphics[width=8.5cm]{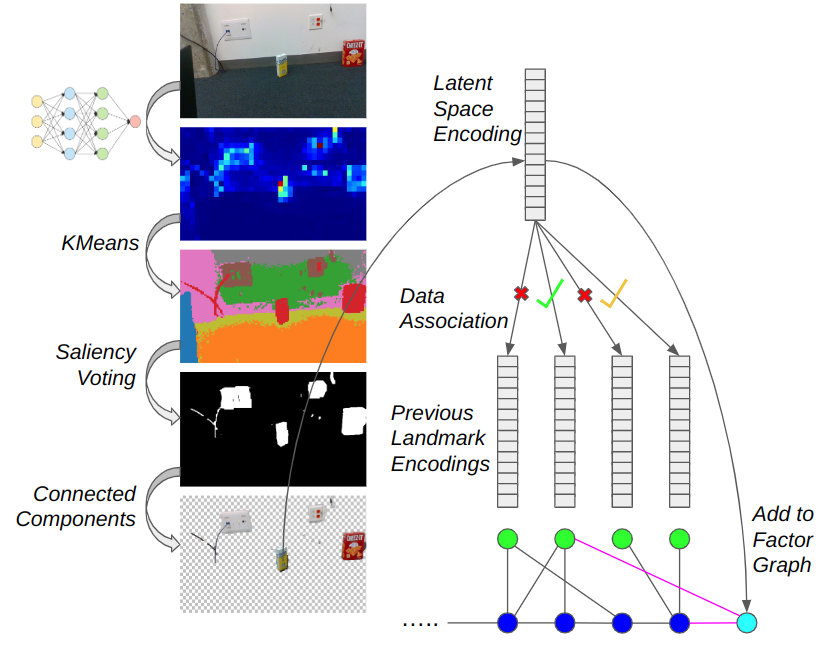}
  \caption{An overview of the proposed open-set data association system coupled with a factor graph framework when a new image and odometry pair is received. The image is fed into the DINO network to get patch-level encodings, which are then clustered into objects. Those clusters are determined to be either foreground or background based on the attention heads. A connected component analysis yields instance level segmentations, from which for each object, a single encoding vector is used as the object representation. The encoding is compared against the existing landmarks' encodings to determine class matches. The pose of the object is also compared against the existing objects' poses as the final data association filter (not pictured). After building a factor with all matches that pass the filter (depending on back-end method, either expectation-maximization, max-mixtures, or max-likelihood factor), we add a new pose (light blue) to the factor graph with a factor connecting it (light pink) to the previous landmark.}
  \label{fig:sys_diagram}
\end{figure}

    \item A tightly-coupled open-set semantic SLAM system that uses the proposed object representation along

    
    with geometric information to improve the vehicle's positioning accuracy and vice-versa

    \item Experimental results on collected and public datasets demonstrating that the proposed method can be used for more accurate and efficient data association and localization compared to dense methods, geometric only methods, and closed-set methods, while also providing more complete maps than closed-set methods.
\end{enumerate}

\section{Related Work}

\subsection{Foundation Models}
A promising avenue for self-supervised segmentation comes from attention-based networks, and especially transformers, which were first proposed by Vaswani et al~\cite{Vaswani2017} for use with natural language processing. Dosovitskiy et al.~\cite{Dosovitskiy2020} extended the use of transformers to image data, demonstrating that they can functionally replace convolutional and recurrent neural networks in a variety of common computer vision tasks. Caron et al.~\cite{Caron2021} showed that self-supervised training with self-attention mechanisms improves the performance of transformers to the level of the previous state-of-the-art for visual feature extraction in a method that they refer to as DINO, with an updated version, DINOv2 proposed by Oquab et al~\cite{oquab2023dinov2}. Finally, Amir et al.~\cite{Amir2021} explore the use of DINO features for semantic segmentation and correspondence matching on common terrestrial objects and scenes. 

Kirillov et al~\cite{kirillov2023segment} utilize a masked autoencoder (MAE) backbone \cite{he2021masked} to encode the image into a latent space, before using an attention-based decoder to generate segmentation masks from a learned encoding of prompts, in contrast to our method based on \cite{Amir2021}, which operates clustering directly on the encoder outputs to obtain segmentations. By doing this, we can obtain latent space object centroids as a byproduct of the clustering process. Our work also differs in that it utilizes a DINO-based backbone for encoding the image, which due to its multi-scale cropping student-teacher approach more explicitly aims for semantically meaningful encodings.

Radford et al~\cite{radford2021learning} proposed Contrastive Language-Image Pre-Training (CLIP) to tie image and text together, providing the possibility for open-set mapping with text labels. Our work focuses solely on lightweight open-set mapping, with the assumption that labels can be added with minimal human interaction or the use of CLIP or something similar such as Grounding DINO \cite{Liu2023} which connects DINO with text labels.

\subsection{Semantic SLAM}
\begin{figure}
\vspace{2mm}
\center
  \includegraphics[width=8cm]{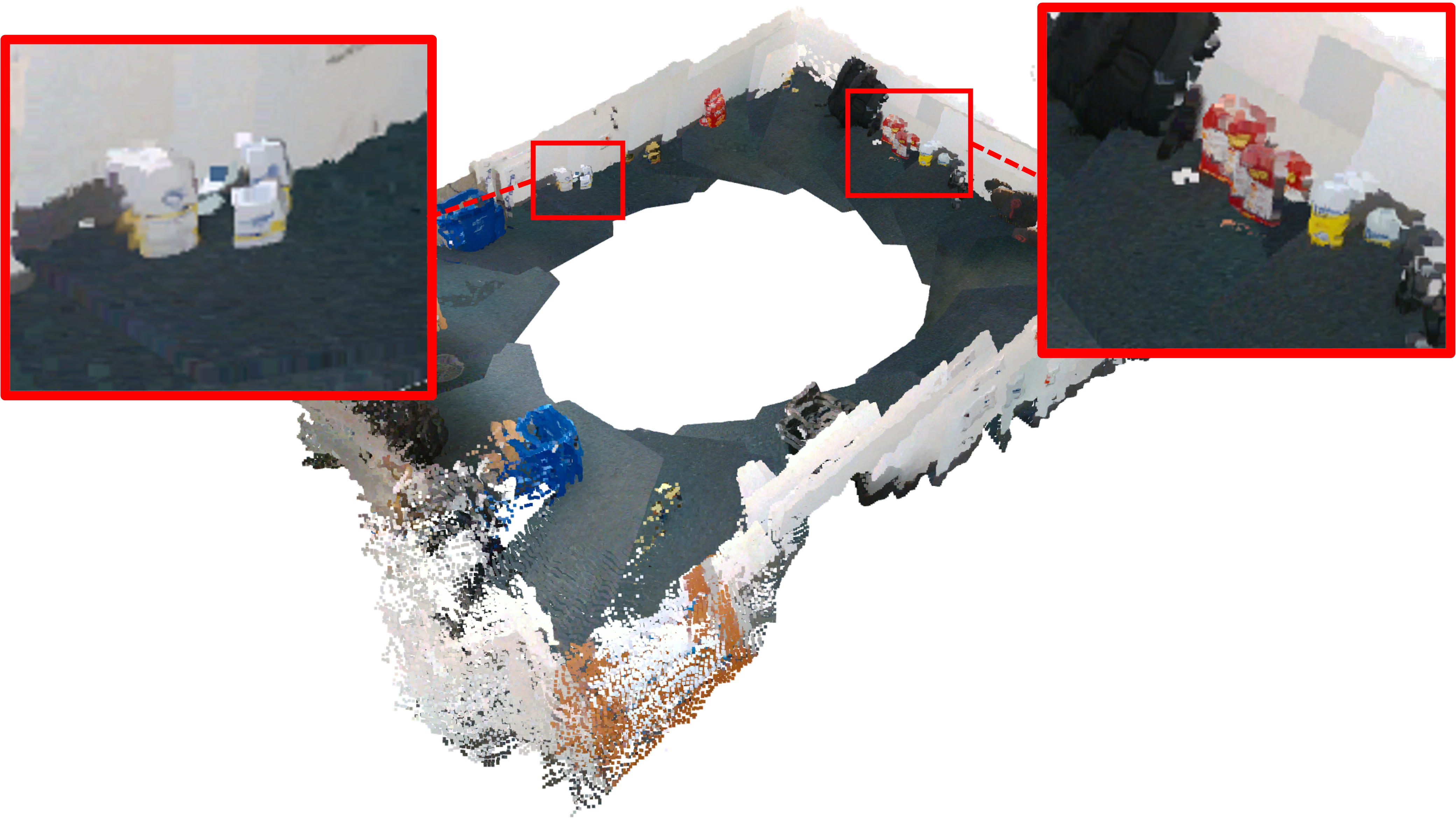}
  \caption{The results of running an existing open-set mapping method \cite{murthy2023} on our data. Artifacts such as the doubled or tripled versions of objects that are a result of not explicitly reasoning about the objects in the scene.}
  \label{fig:concept_fusion}
\end{figure}

Jatavallabhula et al~\cite{murthy2023} propose a method for open-set mapping using both visual-language and visual foundation models. Their method performs dense reconstruction using a $\nabla$-SLAM \cite{Jatavallabhula2019} back-end, unlike our work which is more scalable due to a sparse object representation, and uses a factor graph based back-end representation which allows for the future use of non-differentiable tools for probabilistic data association (in particular discrete-continuous optimization tools \cite{doherty2022discrete}). Their work also uses a K-Means clustering scheme in the feature space to extract semantic segmentations, but due to their dense mapping approach, they do not need any further processing of the segmentation areas, whereas our method refines the segmentations with geometric and heuristic criteria to improve our centroid-based object representation. 

Mazur et al~\cite{Mazur2022} demonstrate a method of neural field feature fusion that performs open-set mapping in real time using an iMAP \cite{sucar2021imap} back-end. Their method is again a dense method that also cannot be easily extended to include efficient multi-hypothesis data association methods. 
Grinvald et al~\cite{Grinvald2019} is another dense semantic mapping method that allows for objects not observed in training to be included in the map. Similar to our method, they run both a semantic segmentation network and geometric segmentation method to refine the semantic methods. However, their geometric method assumes that objects tend to have convex surface geometries, unlike ours. They perform data association through the use of an overlapping area method. Furthermore, they assume localization is known, unlike our method which jointly optimizes over both map and sensor poses. 

Wu et al~\cite{Wu2023} describes a system that includes downstream tasks such as manipulation and active exploration in an object-based mapping framework. They perform data association through finding ORB features within a detected object's bounding box, and then using intersection over union (IOU), single sample-t (S-t), and Neyman-Pearson (NP) tests. A double sample-t test is used to check for duplicates. While they refer to a centroid in their object parameterization (which also includes a semantic label, scale, and position information), the centroid they use is the geometric centroid rather than the latent space centroid used in our work. 
Jamieson et al~\cite{Jamieson2021} demonstrate a method of merging unsupervised, open-set, and therefore inconsistently labeled semantic maps from multiple robots.  

Blomqvist et al~\cite{Blomqvist2023} present a method of open-set semantic segmentation using neural implicit representations.  
Blum et al~\cite{blum2022scim} present perhaps the closest work to ours, in that they use a clustering approach on visual features to perform open-set semantic mapping. However, their focus is on enabling the continuous adaptation of the clustering parameters to improve over the course of a robot's trajectory, and furthermore, their method is also a dense method that grows computationally infeasible over longer periods of time. 
Fu et al~\cite{fu2023neuse} propose a single latent vector embedding for objects that has the advantage of being SE(3)-equivariant, allowing for a object pose estimation and optimization directly in the latent space. However, their method requires a separate set of training data for each object class, making their embedding method infeasible to use for open-set mapping. 
Doherty et al~\cite{Doherty2019} present a closed-set semantic SLAM system with a similar Mahalanobis distance based filtering mechanism for generating hypothesis data associations to ours, albeit with the use of discrete semantic classes rather than a continuous vector space representation. Their work thus utilizes a discrete-continuous optimization framework \cite{doherty2022discrete} to optimize the discrete semantic class assignments as well as the continuous object and vehicle poses, whereas our work simply uses the continuous vector representation of the object as proxy for the object class that can be backed out in post-processing. 
Generally, our work differs from existing works for open-set SLAM in its lightweight, sparse map representation, and the use of that representation for data association that can aid the overall SLAM system. 


\section{Methods}

\subsection{Object Embedding}
The image embedding network, DINO \cite{Caron2021}, follows a student-teacher structure in training, with two separate networks that have the same architecture but different parameters. A different random transformation of an input image $x$ is passed to each network. The student network is denoted $g_{\theta_{s}}$ and the teacher network $g_{\theta_{t}}$. Each network then outputs a feature vector of dimension $k$, which is converted to a normalized probability with a temperature $\tau$ softmax i.e. 

\begin{equation}
    P(x)^{(i)}=\frac{\exp \left(g_{\theta_s}(x)^{(i)} / \tau_s\right)}{\sum_{k=1}^K \exp \left(g_{\theta_s}(x)^{(k)} / \tau_s\right)}
\end{equation}

from which a cross entropy loss

\begin{equation}
\min _{\theta_s} P_t(x) log(P_s(x)))
\end{equation}
is used. This is adapted to be usable with self-supervised learning following the method from \cite{Caron2021} by generating a set $\mathcal{V}$ of global views $x_{1}^g$ and $x_{2}^g$ and multiple, smaller resolution local views. The student sees all views, whereas the teacher only sees the global views, which thus allows the student network to learn a local-to-global correspondence through the following loss 

\begin{equation}
\min _{\theta_s} \sum_{x \in\left\{x_1^g, x_2^g\right\}} \sum_{\substack{x^{\prime} \in V \\ x^{\prime} \neq x}} P_t(x) log( P_s(x^{\prime})).
\end{equation}

The student weights are updated through back-propagation, whereas the teacher network weights are updated as an exponential moving average of the student weights. This network has been demonstrated in previous works \cite{Caron2021} \cite{Amir2021} to produce semantically meaningful features. 

\subsection{Clustering}
We modify the implementation from \cite{Amir2021}, which uses the FAISS library \cite{johnson2019billion} to perform an iterative process of K-Means clustering on the DINO features and obtain semantically segmented areas. The clustering process returns a centroid vector in the latent space (\emph{not} a geometric centroid), which we utilize as our full object representation. Salient clusters are determined through a voting procedure based on the attention heads again following the method from \cite{Amir2021}, which determines which clusters are in the foreground and should be used for mapping. The salient areas are refined through erosion and GrabCut \cite{rother2004grabcut}. Our method to convert these semantic segmentations into instance level segmentations is through a connected components analysis on the salient clusters. We then calculate the geometric centroid of the salient clustered area and find the range to that point in order to obtain a point in space for that object. 
It is not obvious that using the centroid of an object in the latent space will be an effective and meaningful representation of the object that can be used for data association and mapping, since in fact e.g. the ear of a cat and the ear of a dog are more similar in the feature space than the ear of a dog and the tail of a dog \cite{Amir2021}. Thus, one of the contributions of this work is to show experimentally that the latent space centroid can be used effectively as a lightweight representation for the entire object, and that the representation can be used consistently for data association across different viewing angles over the course of trajectory. 

\begin{figure}
\vspace{-2mm}
\end{figure}
\begin{algorithm}[t]
\caption{Instance-Level Open-Set Object Detection and Localization}\label{alg1}
\begin{algorithmic} 
\STATE Given: DINO network $\Theta$
\STATE Inputs: RGBD image $I_{input}$, patch size $n$, saliency threshold $thresh$, minimum object size $min\_size$
\STATE Output: List of object centroid locations and encodings
\FOR{$n$ x $n$ patch $P$ in $I_{input}$}
\item  DINO features $ \leftarrow \Theta(P)$
\ENDFOR
\STATE Clusters $\leftarrow$ K-Means Clustering(DINO features) 
\FOR{C in Clusters}
\STATE Saliency voting using attention head weights  
\IF{$C_{votes}$ $>$ $thresh$}
\STATE Salient Clusters $\leftarrow$ C 
\ENDIF
\ENDFOR 
\FOR{Class in Salient Clusters Classes}
\item Convert salient clusters of Class to binary image $I_{fg/bg}$
\item $I_{fg/bg} \leftarrow Erode(I_{fg/bg}$)
\item $I_{fg/bg} \leftarrow GrabCut(I_{fg/bg}$)
\item $ConnectedComponents(I_{fg/bg}$)
\FOR{component in Connected Components}
\IF{$size(component) < min\_size$}
\STATE Remove from cluster list
\ENDIF
\IF{$component \cap image\_border \neq \phi$}
\STATE Remove from cluster list
\ENDIF
\ENDFOR
\ENDFOR
\STATE Retrieve geometric and feature space centroid for remaining clusters
\STATE Return: Latent space centroids, Geometric centroids

\end{algorithmic}
\end{algorithm}

\subsection{SLAM Framework}
The problem of semantic SLAM is formulated as a \textit{maximum a posteriori} (MAP) problem as follows: 

\begin{equation}
  \label{eq:semantic-slam}
  \Xopt, \Lopt = \argmax_{\poses, \landmarks} p(\poses, \landmarks \mid \measurements).
\end{equation}
where $\poses \triangleq \{\pose_i : \pose_i \in \SE(3), i = 1, \ldots
N\}$ are robot poses, and  $\landmarks \triangleq \{\landmark_j \triangleq (\geometric{\landmark}_j,
\semantic{\landmark}_j),\ j = 1, \ldots, M$\} are semantic landmarks,
where each landmark is split into a continuous geometric component
and a discrete semantic class component. Discrete class assignments are determined through equation~\ref{eq:cossim} and are used only for admitting data association hypotheses. Thus we only need to consider the geometric component of the landmark measurements $\geometric{\landmark}_j \in \SE(3)$. The associations between landmark observations and previously seen landmarks is considered unknown in our work, and thus these data associations must also be determined as explained in section \ref{sec:da}, thus making equation \eqref{eq:semantic-slam} with the inclusion of data association variable \associations

\begin{equation}
  \label{eq:semantic-slam-unknown-da}
  \Xopt, \Lopt, \Dopt = \argmax_{\poses, \landmarks, \associations} p(\poses, \landmarks, \associations \mid \measurements).
\end{equation} 

By marginalizing out the data association variables:

\begin{subequations}
  \begin{equation}
    \Xopt, \Lopt = \argmax_{X, L} \left[  \max_D p(X, L, D \mid Z)  \right], \label{eq:max-marg-opt}
  \end{equation}
  \begin{equation}
    X^{+}, L^{\mathtt{+}} = \argmax_{X, L} \underbrace{\sum_D p(X, L, D \mid Z)}_{p(X,L \mid Z)}, \label{eq:sum-marg-opt}
  \end{equation}
\end{subequations}

where $X^{+}, L^{\mathtt{+}}$ are the marginal MAP estimates (as opposed to MAP) we can formulate the problem as a factor graph with variables for $\poses$ and $\landmarks$, and by leveraging the conditional independence structure of the factor graph we can use iSAM2 \cite{Kaess12ijrr} to obtain MAP estimates.

\subsection{Data Association} \label{sec:da}
We utilize a cosine similarity metric to determine whether objects are of the same class. 

\begin{equation}\label{eq:cossim}
sim_{cos}(A, B) =   \dfrac {A \cdot B} {\left\| A\right\| _{2}\left\| B\right\| _{2}} 
\end{equation}

Given that two objects are of the same class as determined by a cosine similarity threshold $\alpha$

\begin{equation}\label{eq:cossim_thresh}
sim_{cos}(A,B) > \alpha
\end{equation}

we additionally check that the Mahalanobis distance between the observed object and existing object falls under a set threshold. Following the derivation from Kaess et al~\cite{Kaess09ras}, we can calculate the Mahalanobis distance between the landmark measurement $\tilde{\mathbf{z}}_k$ and landmark correspondence hypothesis $j_k=j$ given the state $\mathbf{x}$ and all previous measurements $Z^{-}$ as 

\begin{equation}
\begin{aligned}
    & P(\tilde{\mathbf{z}}_k, j_k=j \mid Z^{-}) \\
    &= \int_{\mathbf{x}} P(\tilde{\mathbf{z}}_k, j_k=j, \mathbf{x} \mid Z^{-}) \\
    &= \int_{\mathbf{x}} P(\tilde{\mathbf{z}}_k, j_k=j \mid \mathbf{x}, Z^{-}) P(\mathbf{x} \mid Z^{-}) \\
    &= \int_{\mathbf{x}} P(\tilde{\mathbf{z}}_k, j_k=j \mid \mathbf{x}) P(\mathbf{x} \mid Z^{-}) \\
    & = \int_{\mathbf{x}} \frac{1}{\sqrt{|2 \pi \Gamma|}} e^{-\frac{1}{2}\left\|h_{i_k j}(\mathbf{x})-\tilde{\mathbf{z}}_k\right\|_{\Gamma}^2} \frac{1}{\sqrt{|2 \pi \Sigma|}} e^{-\frac{1}{2}\|\mathbf{x}-\hat{\mathbf{x}}\|_{\Sigma}^2} \\
    & \approx \frac{1}{\sqrt{\left|2 \pi C_{i_k j}\right|}} e^{-\frac{1}{2}\left\|h_{i_k j}(\hat{\mathbf{x}})-\tilde{\mathbf{z}}_k\right\|_{C_{i_k} j}^2}
\end{aligned}
\end{equation}

\begin{figure}
\vspace{-1mm}
\end{figure}
where $\|\mathbf{x}\|_{\Sigma}^2:=\mathbf{x}^T \Sigma^{-1} \mathbf{x}$, $h$ is the measurement model, and the covariance $C_{i_k j}$ is defined as

$$
C_{i_k j}:=\left.\left.\frac{\partial h_{i_k j}}{\partial \mathbf{x}}\right|_{\hat{\mathbf{x}}} \Sigma \frac{\partial h_{i_k j}}{\partial \mathbf{x}}\right|_{\hat{\mathbf{x}}} ^T+\Gamma .
$$

Thus our distance function is

\begin{equation} \label{eq:ml_distance}
   D_{k j}^{2, \mathrm{ML}}:=\left\|h_{i_k j}(\hat{\mathbf{x}})-\tilde{\mathbf{z}}_k\right\|_{C_{i_k j}}^2 
\end{equation}

where again we evaluate the hypothesis that a specific measurement $\tilde{\mathbf{z}}_k$ taken in image $i_k$ was caused by the $j^{t h}$ landmark. Equation~\ref{eq:ml_distance} follows a chi-squared distribution, and thus we can use a $d$-degree of freedom chi-square test:

\begin{equation}\label{eq:chi_square}
D_{k j}^{2, \mathrm{ML}}<\chi_{d, \beta}^2
\end{equation}

If the observed object meets both the criteria (equation~\ref{eq:chi_square} and equation~\ref{eq:cossim_thresh}), we add the existing observation to the list of hypothesis data associations. 

After checking against all existing objects within a given geometric distance, we rank the hypotheses based on the log marginal measurement likelihood.
For max-likelihood, we simply select the most likely estimate given prior estimates $\poses^0, \landmarks^0$ as the associated landmark for the observation, and then use that fixed data association for computing landmark and pose estimates:

\begin{equation}
\begin{gathered}
\hat{\associations}=\underset{\associations}{\arg \max } \; p\left(\associations \mid \poses^0, \landmarks^0, \measurements\right) \\
\hat{\poses}, \hat{\landmarks}=\underset{\poses, \landmarks}{\arg \max } \log p(\measurements \mid \poses, \landmarks, \hat{\associations})
\end{gathered}
\end{equation}

For max-mixtures, we build a mixture factor as in \cite{Doherty2019}, where we can marginalize out the data association factor given previous observations $\measurements^{-}$ and new measurements $\measurements^{+}$: 

\begin{multline}
  \hat{p}(\poses, \landmarks \mid \measurements^{+}, \measurements^{-}) = \\
  p(\poses, \landmarks \mid \measurements^{-}) \max_{\associations^{+}} \left[ p(\measurements^{+} \mid \poses, \landmarks, \associations^{+}) p(\associations^{+} \mid \measurements^{-}) \right].
\end{multline}

For expectation maximization, we calculate the marginal MAP as in \cite{bowman2017probabilistic} where $\poses^i, \mathcal{L}^i$ are initial estimates:

$$
\begin{aligned}
X^{+}, & \; L^{\mathtt{+}} =\underset{\poses, \landmarks}{\arg \max } \; \mathbb{E}_{\associations}\left[\log p(\measurements \mid \poses, \landmarks, \associations) \mid \poses^i, \landmarks^i, \measurements\right] \\
& =\underset{\poses, \landmarks}{\arg \max } \sum_{\associations \in \mathbb{D}} p\left(\associations \mid \poses^i, \landmarks^i, \measurements\right) \log p(\measurements \mid \poses, \landmarks, \associations)
\end{aligned}
$$
where $\mathbb{D}$ is the space of all possible values of $\mathcal{D}$. 

If there are no hypotheses that meet both of the criteria, the observation is determined to be of a new landmark, and a new landmark observation is added to the optimization. 



\section{Experimental Results}

\subsection{Datasets}
Experimental data was collected using an Intel RealSense D435i for RGBD data and an OptiTrack motion capture system for ground truth trajectories. A variety of objects were placed along the floor, and a Clearpath Jackal was navigated through three loops around the room with the RealSense mounted onboard. 
A second sequence of three loops was collected with objects of the same class viewed from different angles. 

We also evaluate our method on the publicly available TUM Pioneer Robot datasets \cite{sturm12iros} that include a much larger scene, for which we use the provided wheel odometry. 


\subsection{Alternate Methods Comparison}
We compare our method with a popular existing open-set semantic mapping system, ConceptFusion \cite{murthy2023}. ConceptFusion uses all available GPU memory quickly, and thus comparisons were made to a modified version of their implementation where the number of images used was downsampled to 1/70th of the total number of images processed by our system. We also provided their system with the same noisy odometry as our method since their built-in odometry system relies on frame-to-frame alignments that are not possible at subsampled framerates. Qualitative results for ConceptFusion are shown in \ref{fig:concept_fusion}. A quantitative evaluation of their trajectory was not possible as they do not jointly optimize their trajectory with the observed objects. 

We also run a comparison against a modern closed-set detector (YOLOv8~\cite{Jocher_YOLO_by_Ultralytics_2023}) whose class is then processed as a one-hot encoding which can be fed into the rest of our system. 
As an ablation, we run a method that chooses the most likely data association based only on Mahalanobis distance without using the landmark encoding cosine similarity metric which we refer to in results tables as Geometric Only. We provide a trajectory from ORBSLAM3 as a feature-based comparison, with results similar to our noise multiplier 4 accuracy. For our method, geomtric only, and ConceptFusion, odometry was provided by the ground truth pose with added Gaussian noise. A systematic study of the effects of increasing the noise on the results of the system is presented in Table~\ref{table:mocapseq1} and Table~\ref{table:mocapseq2}. Results demonstrate that our method is able to successfully mitigate odometry-based drift through the use of open-set object detections, and furthermore, that the use of the object encodings aids the accuracy of the trajectory. 
Trajectories were evaluated against the ground truth using evo \cite{grupp2017evo}. 

\subsection{Compute Resource Usage}
\begin{figure}
\center
  \includegraphics[width=.9\linewidth]{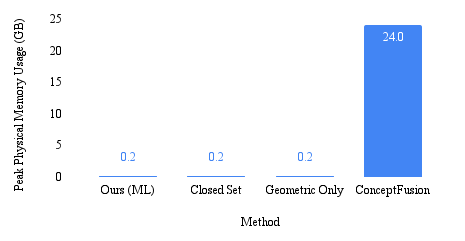}
  \caption{Memory usage on sequence 1 of collected data. Sparse methods drastically reduce the memory consumption for open-set SLAM as compared to dense methods. }
  \label{fig:mem_usage}
\end{figure}

DINO feature extraction is completed as a preprocessing step as done in competing methods, at a rate 4.6 seconds per image on an NVIDIA GeForce RTX 3080 Ti Laptop GPU. The rest of our system runs in real time at 5 Hz, limited by the number of images available rather than compute resources. Due to the sparse representation and the ability to close loops, the memory and storage usage of our system grows at a much slower rate than competing dense methods, while we simultaneously achieve accurate trajectories. A comparison of peak memory usage is presented in Figure~\ref{fig:mem_usage}. 

\begin{figure}[h] 
\center
  \includegraphics[width=8cm]{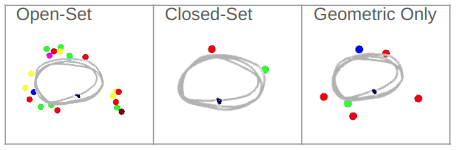}
  \caption{The closed-set map has a less accurate trajectory, and also identifies fewer objects in the scene. The geometric only data association method incorrectly associates objects in close proximity as being the same object as it does not have the object encoding to help differentiate those instances; the incorrect data associations result in a less accurate map and trajectory.\\
  Object color to class mappings were identified by a human in post-processing as follows for open-set: Red - electric socket; Green - sugar box; Dark blue - CheezIt box; Yellow - spam can; Pink - skateboard; White - trash bag; Light blue - trash bin. For closed-set: Red - skateboard; Green - trash can. For geometric only (note that the classes were not used during mapping, and are identified for comparison purposes only): Red - electric socket; Green - sugar box; Dark blue - Skateboard.}
  \label{fig:traj_and_lms}
\end{figure}

\begin{figure}
     \centering
     \begin{subfigure}[b]{0.48\linewidth}
     \center
         \includegraphics[width=.9\linewidth]{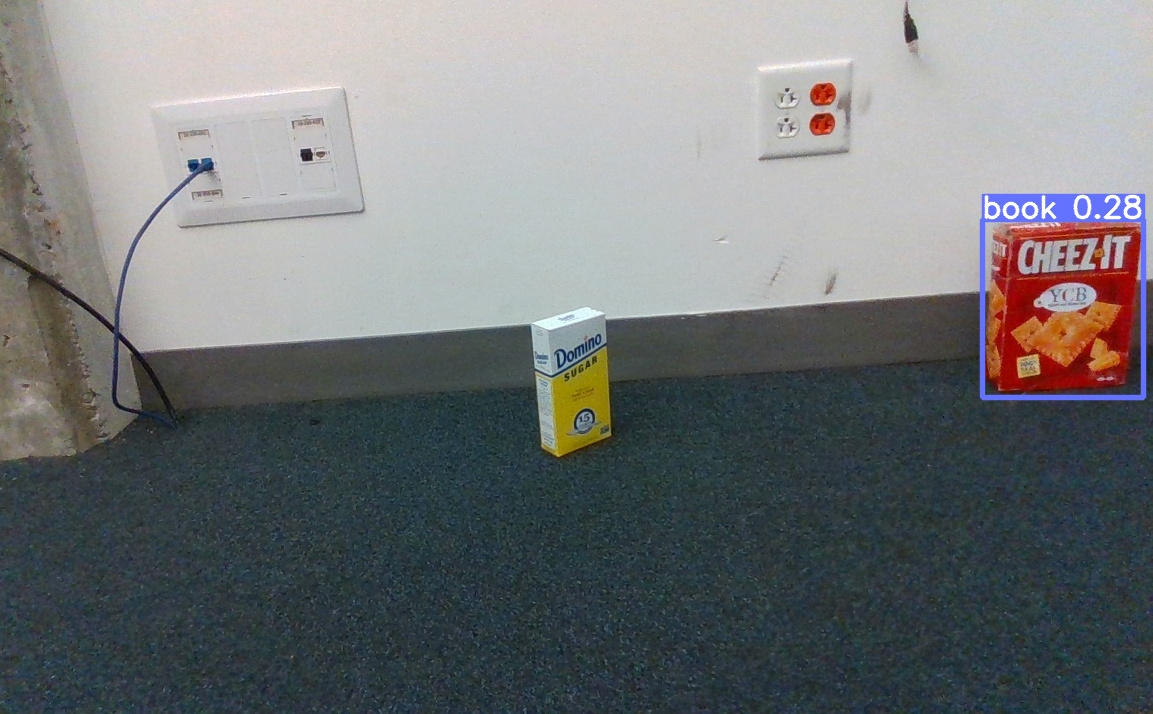}
         \caption{The detector incorrectly identifies the Cheez-It box as a book, and fails to identify the other three objects in the scene. }
         \label{fig:yolo_cheezit}
     \end{subfigure}
     \hfill
     \begin{subfigure}[b]{0.48\linewidth}
     \center
         \includegraphics[width=1\linewidth]{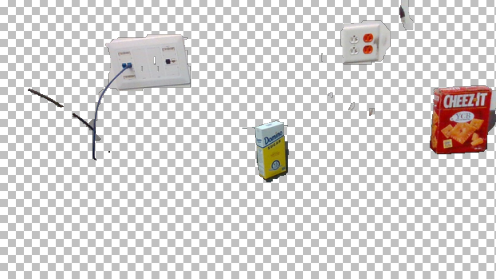}
         \caption{An open-set detector is able to extract all objects in the scene and our method assigns each one an encoding vector.}
         \label{fig:ours_cheezit}
     \end{subfigure}
        \caption{Images from the collected data were fed to YOLOv8 \cite{Jocher_YOLO_by_Ultralytics_2023}, a state-of-the-art and widely used object detector. The detector failed to identify many of the common everyday items in the scene, and even incorrectly labeled one item. Our open-set detector identifies each object and associates it with a latent vector encoding.}
        \label{fig:yolo_performance}
\end{figure}

\begin{figure*}[h]
    \centering
    
    \begin{subfigure}[t]{0.3\linewidth}
        \includegraphics[width=\linewidth]{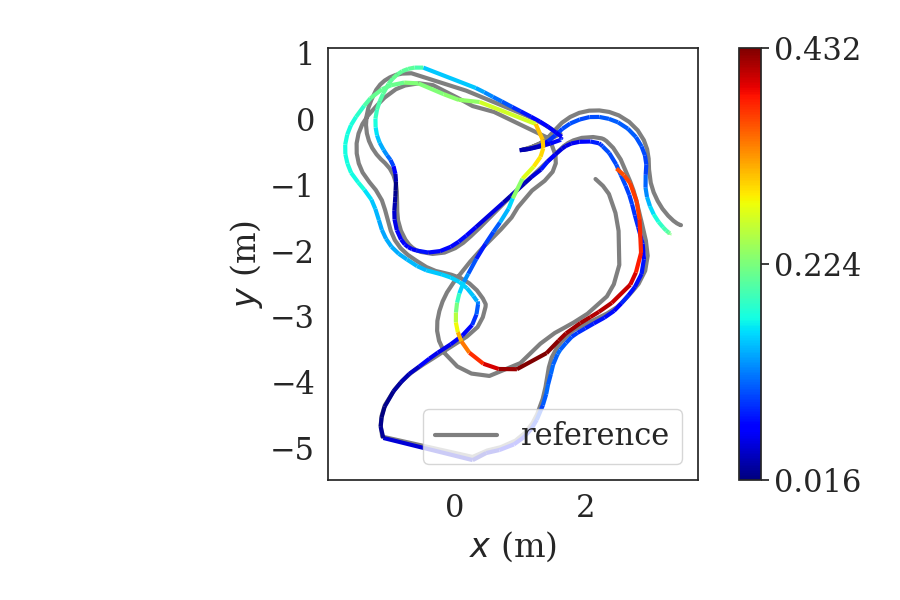}
        \caption{Open-set maximum likelihood trajectory for TUM Pioneer SLAM dataset colored by APE w.r.t translation (m)}
        \label{fig:sub1}
    \end{subfigure}
    \hfill
    \begin{subfigure}[t]{0.3\linewidth}
        \includegraphics[width=\linewidth]{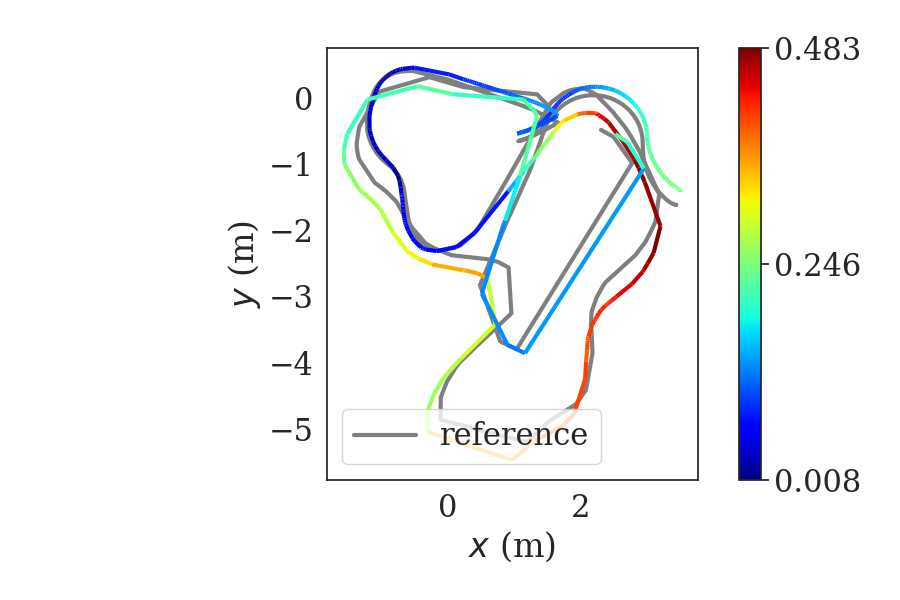}
        \caption{Open-set expectation maximization trajectory for TUM Pioneer SLAM dataset colored by APE w.r.t translation (m)}
        \label{fig:sub2}
    \end{subfigure}
    \hfill
    \begin{subfigure}[t]{0.3\linewidth}
        \includegraphics[width=\linewidth]{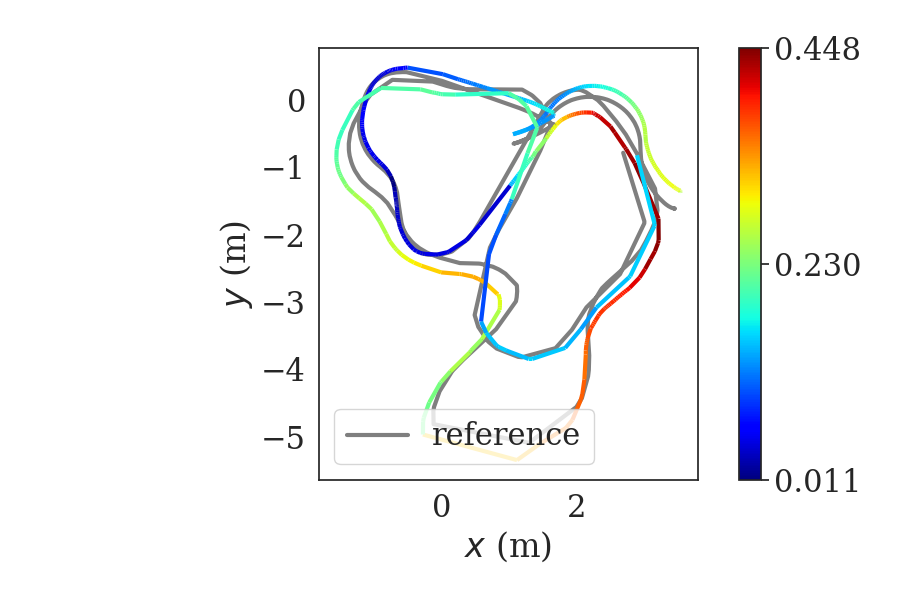}
        \caption{Open-set max-mixtures trajectory for TUM Pioneer SLAM dataset colored by APE w.r.t translation (m)}
        \label{fig:sub3}
    \end{subfigure}
    \medskip
    \begin{subfigure}[t]{0.3\linewidth}
        \includegraphics[width=\linewidth]{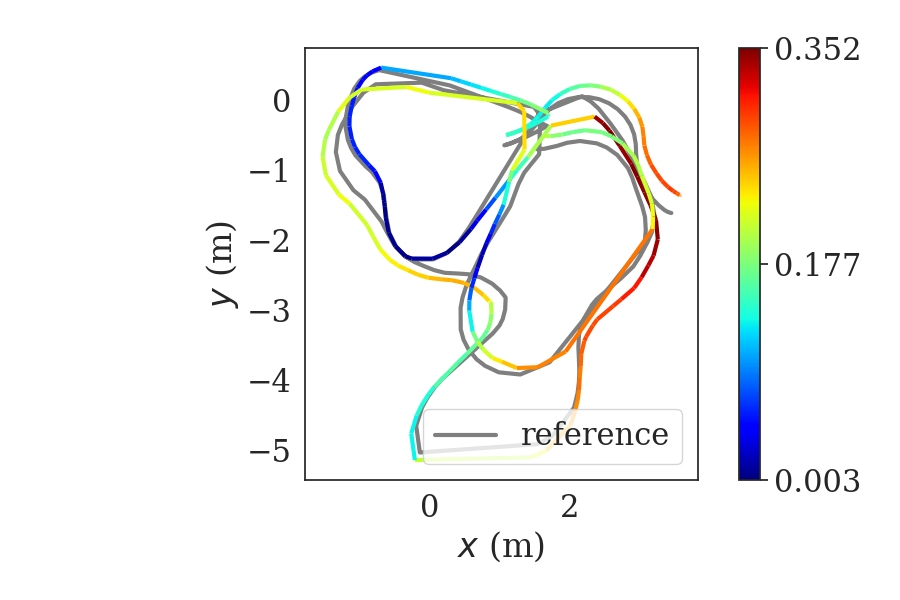}
        \caption{Closed-set max-likelihood trajectory for TUM Pioneer SLAM dataset colored by APE w.r.t translation (m)}
        \label{fig:sub4}
    \end{subfigure}
    \hfill
    \begin{subfigure}[t]{0.3\linewidth}
        \includegraphics[width=\linewidth]{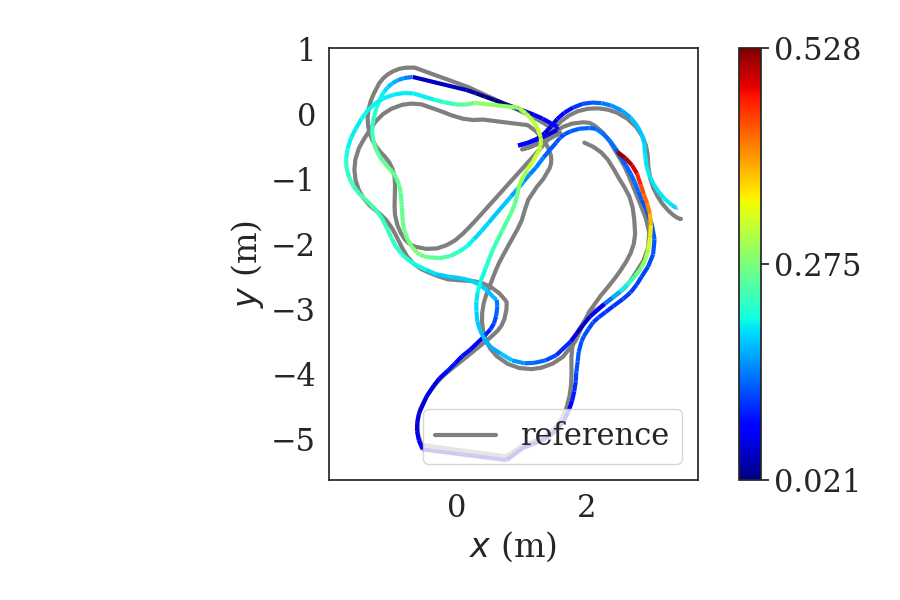}
        \caption{Geometric only max-likelihood trajectory for TUM Pioneer SLAM dataset colored by APE w.r.t translation (m)}
        \label{fig:sub5}
    \end{subfigure}
    \hfill
    \begin{subfigure}[t]{0.3\linewidth}
        \includegraphics[width=\linewidth]{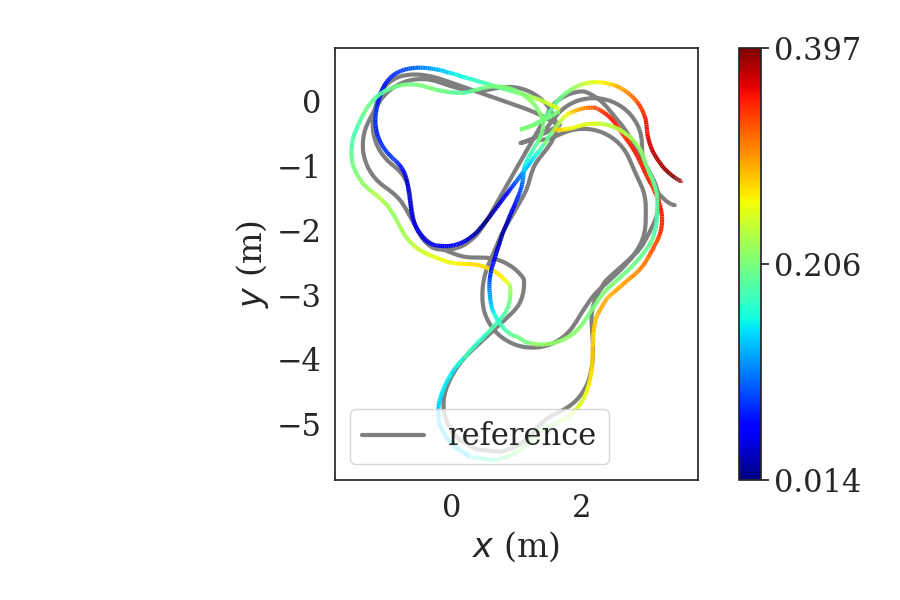}
        \caption{Wheel odometry trajectory for TUM Pioneer SLAM dataset colored by APE w.r.t translation (m)}
        \label{fig:sub6}
    \end{subfigure}
    
    \caption{Our method performs best regardless of data association method. Maximum likelihood has the best performance for open-set, closed-set, and geometric only, and thus for closed-set and geometric-only, we only show results for maximum likelihood data association.}
    \label{fig:multisubfigures}
\end{figure*}


\section{Discussion}
It is interesting to note that by enabling the system to run open-set segmentation, several unanticipated objects were identified and used by the system. These include electric sockets, Ethernet sockets, and large scuffs on the wall. 

The system is sensitive to clustering hyper-parameters and saliency thresholds, as for example the bag inside the trash can was consistently identified, while the can itself was not. By modifying these parameters, it is possible to change the granularity of the detected objects. 

While occlusions are not explicitly handled, sections of the trajectory do include partial and occluded views of objects, and inter-object occlusion results in negligible performance degradation of our method. Occlusions remain a challenge even for existing methods since bounding box centroids and other common geometric representations encounter the same issue of inconsistently identified geometric centroids. 

While the closed-set method achieves lower pose error in sequence 2, the open-set methods have only slightly higher error in this sequence, while also building a more complete map. 

The max-mixture method for data association consistently performed worse than max-likelihood and expectation-maximization, which is most likely due to the non-convexity of the cost landscape that is introduced by the method, which thus makes it difficult for the local optimization methods used in GTSAM to move to the global minimums. 

\section{Conclusion}
In conclusion, we present a novel system for tightly-coupled open-set semantic SLAM in sparse environments. We take an off-the-shelf image encoding network and run several post-processing steps to obtain instance-level object segmentations. We propose a lightweight single vector encoding for each object, and demonstrate that the object encoding is amenable to several data association methods in a factor graph-based SLAM framework. Our method is computationally more efficient than competing dense methods, and achieves high localization accuracy. By reasoning at the object level rather than at the pixel or dense feature level, the maps from our method are more semantically consistent than dense methods. Our method builds a more complete map and achieves higher localization accuracy than closed-set methods. The code and data is open-sourced for the robotics community.

\begin{table*}
\vspace{6mm}
\begin{center}
\caption{Sequence 1, Average pose error in meters: Gaussian noise with base sigmas of $(\sigma^2_{x}, \sigma^2_{y}, \sigma^2_{z}, \sigma^2_{r}, \sigma^2_{p}, \sigma^2_{y})$ = (0.001, 0.001, 0.001, 0.001, 0.001, 0.001) was added to the relative odometry measurements at each of the 1,838 keyframes, which occurred at 5 Hz. The indicated added noise multipliers were multiplied with the base sigmas. The entire trajectory was 21.7 meters. The ORBSLAM3 result is included without added noise as a baseline comparison. Only the best performing data association method result is included for closed-set and geometric only.}
\begin{tabular}{c c c c c c c} 
 \hline
 Added noise multiplier & Ours (ML) & Ours (EM) & Ours (MM) & Closed-Set (MM) & Geometric Only & Noisy Odometry \\ [0.1ex] 
 \hline
  
  1 & 0.021 & 0.011 & 0.014 & \textbf{0.007} & 0.444 & 0.035 \\ 
 
 2 & 0.039 & 0.022 & 0.029 & \textbf{0.019} &  0.441 & 0.047\\
 
 3 & 0.057 & \textbf{0.024} & 0.044 & 0.032 & 0.436 & 0.061\\
 
 4 & 0.073 & \textbf{0.031} & 0.058 & 0.041 & 0.450 & 0.081\\
 
 5 & 0.095 & \textbf{0.038} & 0.044 & 0.049 & 0.449 & 0.098\\ 
 \hline 
 ORBSLAM3 (Baseline) & 1.490 \\[1ex] 
 
\end{tabular}

\label{table:mocapseq1}
\end{center}
\vspace{-5mm}

\end{table*}

\begin{table*}[]
\begin{center}
\caption{Sequence 2, Average pose error in meters: This sequence contains objects of the same class but with more varied viewing angles demonstrating that our data association scheme is robust to viewing angle changes of the same class of objects. Gaussian noise with base sigmas of $(\sigma^2_{x}, \sigma^2_{y}, \sigma^2_{z}, \sigma^2_{r}, \sigma^2_{p}, \sigma^2_{y})$ = (0.001, 0.001, 0.001, 0.001, 0.001, 0.001) was added to the relative odometry measurements at each of the 1,642 keyframes, which occurred at 5 Hz. The indicated added noise multipliers were multiplied with the base sigmas. The entire trajectory was 31.7 meters. The ORBSLAM3 result is included without added noise as a baseline comparison. Only the best performing data association method result is included for closed-set and geometric only.}
\begin{tabular}{c c c c c c c} 
 \hline
  Added noise multiplier & Ours (ML) & Ours (EM) & Ours (MM) & Closed-Set (MM) & Geometric Only & Noisy Odometry \\ [0.5ex] 
 \hline
 1 & 0.014 & 0.008 & 0.011 & \textbf{0.007} & 0.137 & 0.053\\ 
 2 & 0.024 & 0.016 & 0.023 & \textbf{0.018} & 0.132 & 0.064\\
 3 & 0.036 & 0.031 & 0.030 & \textbf{0.023} & 0.141 & 0.069\\
 4 & 0.048 & 0.031 & 0.042 & \textbf{0.023} & 0.150 & 0.079\\
 5 & 0.061 & 0.039 & 0.046 & \textbf{0.037} &  0.136 & 0.097 \\ 
 \hline
 ORBSLAM3 (Baseline) & 0.041 \\[1ex] 
\end{tabular}
\label{table:mocapseq2}
\end{center}
\vspace{-5mm}

\end{table*}

\begin{table*}[]
\begin{center}
\caption{TUM Pioneer Datasets results, Average Pose Error in meters. ML: Max-likelihood, EM: Expectation-Maximization, MM: Max-Mixtures. Only the best performing data association method result is included for closed-set and geometric only.}
\begin{tabular}{c c c c c c c} 
 \hline
 Dataset & Ours (ML) & Ours (EM) & Ours (MM) & Closed-Set (ML) & Geometric Only (ML) & Wheel odometry only\\ [0.1ex] 
 \hline 
  Pioneer SLAM & \textbf{0.139} & 0.184 & 0.197 & 0.180 & 0.180 & 0.202 \\ 
 
 Pioneer SLAM 2 & \textbf{0.231} & 0.342 & 0.341 & 0.602 & 0.472 & 0.341 \\
 [1ex] 
\vspace{-5mm}

\end{tabular}

\label{table:mocapseq1}
\end{center}
\vspace{-5mm}
\end{table*}

\section*{Acknowledgments}
\footnotesize This work was supported by the MIT Lincoln Laboratory Autonomous Systems Line which is funded by the Under Secretary of Defense for Research and Engineering through Air Force Contract No. FA8702-15-D-0001, ONR grants
N00014-18-1-2832, N00014-23-12164, and N00014-19-1-2571 (Neuroautonomy MURI), and the MIT Portugal Program.

\bibliographystyle{IEEEtran}
\bibliography{main}

\end{document}